\documentclass[letterpaper, 10 pt, conference]{ieeeconf}  

\IEEEoverridecommandlockouts                              
\usepackage{graphicx}
\usepackage{amsmath}
\usepackage{amssymb}
\usepackage{subcaption}
\usepackage{color,soul}
\usepackage{epsfig}
\usepackage{amsmath}
\usepackage{amssymb}
\usepackage{algorithm}

\usepackage{makecell}
\usepackage{algpseudocode}
\usepackage{subcaption}
\usepackage{multirow}
\usepackage{color}
\usepackage{multicol}
\usepackage[colorinlistoftodos,prependcaption,textsize=tiny]{todonotes}
\usepackage{marginnote}
\usepackage{subcaption}
\usepackage{pifont}
\usepackage{url}  
\usepackage{graphicx}  
\usepackage[normalem]{ulem}

\title{\LARGE \bf
Video Object Segmentation using Teacher-Student Adaptation \\ in a Human Robot Interaction (HRI) Setting
}

\author{ Mennatullah Siam$^{1}$, Chen Jiang$^{1}$, Steven Lu$^{1}$, Laura Petrich$^{1}$, \\ Mahmoud Gamal$^{2}$, 
Mohamed Elhoseiny$^{3}$, Martin Jagersand$^{1}$ 
\thanks{$^{1}$Mennatullah Siam, Chen Jian, Steven Lu, Laura Petrich and Martin Jagersand are with the University of Alberta, Canada. 
e-mail: mennatul@ualberta.ca.}
\thanks{$^{2}$Mahmoud Gamal is with Cairo University, Egypt.}
\thanks{$^{3}$ Mohamed El-Hoseiny is with Facebook AI Research.
e-mail: elhoseiny@fb.com.}
}

\begin{document}

\maketitle
\thispagestyle{empty}
\pagestyle{empty}

\begin{abstract}
Video object segmentation is an essential task in robot manipulation to facilitate grasping and learning affordances. Incremental learning is important for robotics in unstructured environments. Inspired by the children learning process, human robot interaction (HRI) can be utilized to teach robots about the world guided by humans similar to how children learn from a parent or a teacher. A human teacher can show potential objects of interest to the robot, which is able to self adapt to the teaching signal without providing manual segmentation labels. We propose a novel teacher-student learning paradigm to teach robots about their surrounding environment. A two-stream motion and appearance "teacher" network provides pseudo-labels to adapt an appearance "student" network. The student network is able to segment the newly learned objects in other scenes, whether they are static or in motion. We also introduce a carefully designed dataset that serves the proposed HRI setup, denoted as (I)nteractive (V)ideo (O)bject (S)egmentation. Our IVOS dataset contains teaching videos of different objects, and manipulation tasks. Our proposed adaptation method outperforms the state-of-the-art on DAVIS and FBMS with 6.8\% and 1.2\% in F-measure respectively. It improves over the baseline on IVOS dataset with 46.1\% and 25.9\% in mIoU.
\end{abstract}

\section{Introduction}
The robotics and vision communities greatly improved video object segmentation over the recent years. The main approaches in video object segmentation could be categorized into semi-supervised and unsupervised approaches. In semi-supervised video object segmentation approaches (e.g., \cite{voigtlaender2017online}\cite{caelles2016one}\cite{khoreva2016learning}, the method is initialized manually by a segmentation mask in the first few frames, then the segmented object is tracked throughout the video sequence. On the other hand, unsupervised methods \cite{kohprimary}\cite{tokmakov2017learning}\cite{jain2017fusionseg}\cite{tokmakov2016learning} attempt to discover the primary object automatically and segment it through the video sequence. Motion is one of the fundamental cues that can help improve unsupervised video object segmentation. While there has been recent success in deep learning approaches for segmenting motion (e.g., \cite{tokmakov2017learning}\cite{jain2017fusionseg}\cite{tokmakov2016learning}), current approaches depend mainly on prior large-scale training data.

\begin{figure}[]
\centering
    \includegraphics[scale=0.35]{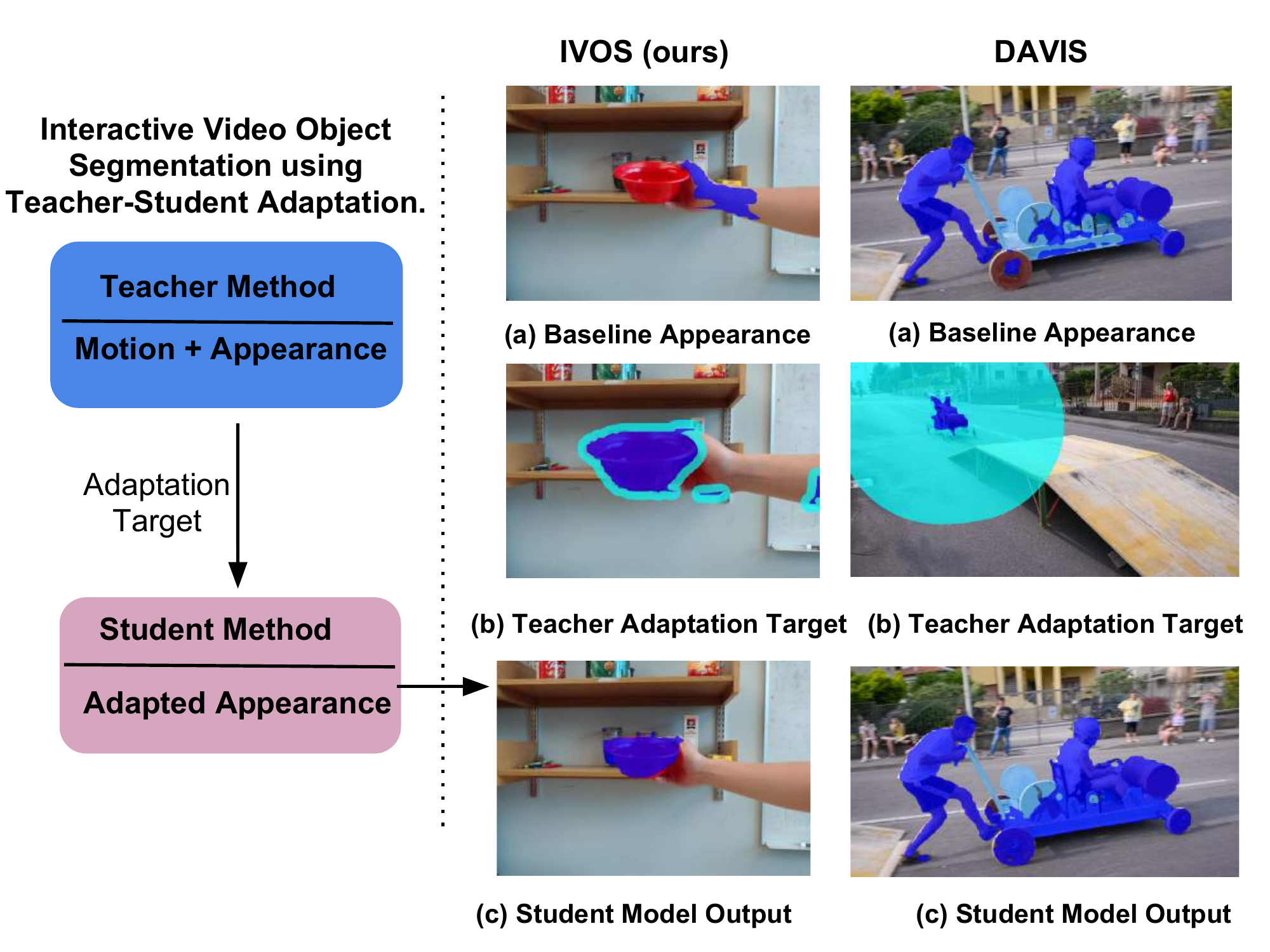}
    \caption{Overview of the proposed Teacher-Student adaptation method for video object segmentation. The teacher model based on motion cues is able to provide pseudo-labels to adapt the student model. Blue: confident positive pixels. Cyan: ignored region in the adaptation.}
     \label{fig:overview}
\end{figure}

Video semantic segmentation for robotics is widely used in different applications such as autonomous driving \cite{cordts2016cityscapes}\cite{ros2016synthia}, and robot manipulation \cite{do2017affordancenet}\cite{kenney2009interactive}. Object segmentation can aid in grasping, manipulating objects, and learning object affordances \cite{do2017affordancenet}. In robot manipulation, learning to segment new objects incrementally, has significant importance. Real world environments have far more objects and more appearance variation than can be feasibly trained a-priori. Current large-scale datasets such as Image-Net \cite{krizhevsky2012imagenet} do not cover this. 

A recent trend in robotics is toward human-centered artificial intelligence. Human-centered AI involves learning by instruction using a human teacher. Such human-robot interaction (HRI) mimics children being taught novel concepts from few examples~\cite{markman1989categorization}. In the robotic setting, a human teacher demonstrates an object by moving it and showing different poses, while verbally or textually teaching its label. The robot is then required to segment the objects in other settings where it is either static or manipulated by the human or the robot itself. We demonstrated this HRI setting in our team submission to the KUKA Innovation Challenge at the Hannover Fair~\cite{KUKAIA}. This HRI setting has few differences to conventional video object segmentation: (1) Abundance of the different poses of the object. (2) The existence of different instances/classes within the same category. (3) Different challenges introduced by cluttered backgrounds, different rigid and non-rigid transformations, occlusions and illumination changes. In this paper, we focus on these robotics challenges and provide a new dataset and a new method to study such a scenario.

We collected a new dataset to benchmark (I)nteractive (V)ideo (O)bject (S)egmentation in the HRI scenario. The dataset contains two types of videos: (1) A human teacher showing different household objects in varying poses for interactive learning. (2) Videos of the same objects used in a kitchen setting while serving and eating food. The objects occur both as static objects and active objects being manipulated. Manipulation was performed by both humans and robots. The aim of this dataset is to facilitate incremental learning and immediate use in a collaborative human-robot environments, such as assistive robot manipulation. Datasets that have a similar setting such as ICUBWorld transformations dataset \cite{pasquale2016object}, and the Core50 dataset \cite{lomonaco2017core50} were proposed. These datasets include different instances within the same category. They benchmark solutions to object recognition in a similar HRI setting but do not provide segmentation annotations unlike our dataset. Other datasets were concerned with the activities of daily living such as the ADL dataset \cite{pirsiavash2012detecting}. The dataset was comprised of ego-centric videos for activities. However, such ADL datasets do not contain the required teaching videos to match the HRI setting we are focusing on. Table \ref{table:datasets} summarizes the most relevant datasets suited to the HRI setting.

The main contribution of our collected IVOS dataset is providing the manipulation tasks setting with objects being manipulated by humans or a robot. In addition to providing segmentation annotation for both teaching videos and manipulation tasks. It enables researchers to analyze the effect of different transformations such as translation, scale, and rotation on the incremental learning of video object segmentation. It acts as a benchmark for interactive video object segmentation in the HRI setting. It also provides videos of both human and similarly robot manipulation tasks with the segmentation annotations along with the corresponding robot trajectories. Thus, it enables further research in learning robot trajectories from visual cues with semantics. 

We propose a novel teacher-student adaptation method based on motion cues for video object segmentation. Our method enables a human teacher to demonstrate objects moving with different transformations and associates them with labels. During inference, our approach can learn to segment the object without manual segmentation annotation. The teacher model is a fully convolutional network that combines motion and appearance, denoted as ``Motion+Appearance``. The adapted student model is a one-stream appearance-only fully convolutional network denoted as ``Appearance``. Combining motion and appearance in the teacher network allows the creation of pseudo-labels for adapting the student network. Our work is inspired from the semi-supervised on-line method \cite{voigtlaender2017online}. This work uses manual segmentation masks for initialization. Instead, our approach tackles a more challenging problem and does not require manual segmentation; it relies on the pseudo-labels provided by the teacher model. Figure \ref{fig:overview} shows an overview of the proposed method. The two main reasons behind using the adaptation targets from the teacher model is: (1) The student model is more computationally efficient. The inference and adaptation time for the teacher model is 1.5x of the student model's. The adaptation occurs only once on the first frame, then the more efficient student model can be used for inference. (2) The teacher model can be used to generate pseudo-labels for the potential object of interest. It does not Urequire the human to provide manual segmentation mask during the teaching phase which provides a natural interface to the human. Consequently, the adapted student model can segment the object of the interest whether it is static or moving. If the adapted model was still dependant on optical flow it will only be able to recognize the object in motion. 

Our proposed method outperforms the state-of-the-art on the popular DAVIS \cite{Perazzi2016} and FBMS \cite{ochs2014segmentation} benchmarks with 6.8\% and 1.2\% in F-measure respectively. On our new IVOS dataset results show the motion adapted network outperforms the baseline with 46.1\% and 25.9\% in mIoU on Scale/Rotation and Manipulation Tasks respectively. Our code \footnote{\url{https://github.com/MSiam/motion_adaptation}} and IVOS dataset \footnote{\url{https://msiam.github.io/ivos/}} are publicly available. A video description and demonstration is available at \footnote{\url{https://youtu.be/36hMbAs8e0c}}. Our main contributions are :
\begin{itemize}
\item Providing a Dataset for Interactive Video Object Segmentation (IVOS) in a Human-Robot Interaction setting, and including manipulation tasks unlike previous datasets.
\item A teacher-student adaptation method is proposed to learn new objects from a human teacher without providing manual segmentation labels. We propose a novel pseudo-label adaptation based on a teacher model that is dependant on motion. Adaptation with discrete and continuous pseudo-labels are evaluated to demonstrate different adaptation methods.
\end{itemize}

\newcommand{\xmark}{\ding{55}}%
\newcommand{\cmark}{\ding{51}}
\begin{table}[]
\centering
\caption{Comparison of different datasets. T:Turntable, H:handheld}
\label{table:datasets}
\begin{tabular}{|l|c|c|c|c|c|c|}
\hline
Dataset  & Sess. & Cat. & Obj. & Acq. & Tasks & Seg. \\\hline
RGB-D \cite{schwarz2015rgb} & - & 51 & 300 & T & \xmark & \xmark \\
BIG BIRD \cite{singh2014bigbird} & - & - & 100 & T & \xmark & \xmark \\
ICUB 28 \cite{pasquale2015teaching} & 4 & 7 & 28 & H & \xmark & \xmark \\
ICUB World \cite{pasquale2016object} & 6 & 20 & 200 & H & \xmark & \xmark \\
Core50 \cite{lomonaco2017core50} & 11 & 10 & 50 & H & \xmark & \xmark \\\hline
\textbf{IVOS} & 12 & 12 & 36 & H & \textbf{\cmark} & \textbf{\cmark}\\\hline
\end{tabular}
\end{table}


\section{IVOS Dataset}
\begin{figure*}[ht!]
\centering
\begin{subfigure}{.12\textwidth}
    \includegraphics[scale= 0.1]{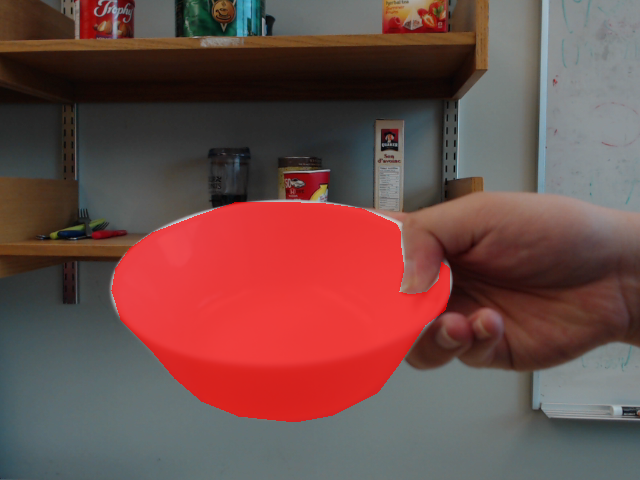}
\end{subfigure}%
\begin{subfigure}{.12\textwidth}
    \includegraphics[scale= 0.1]{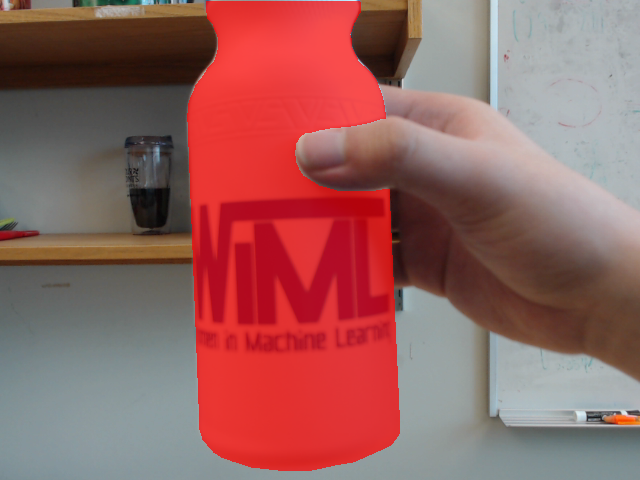}
\end{subfigure}%
\begin{subfigure}{.12\textwidth}
    \includegraphics[scale= 0.1]{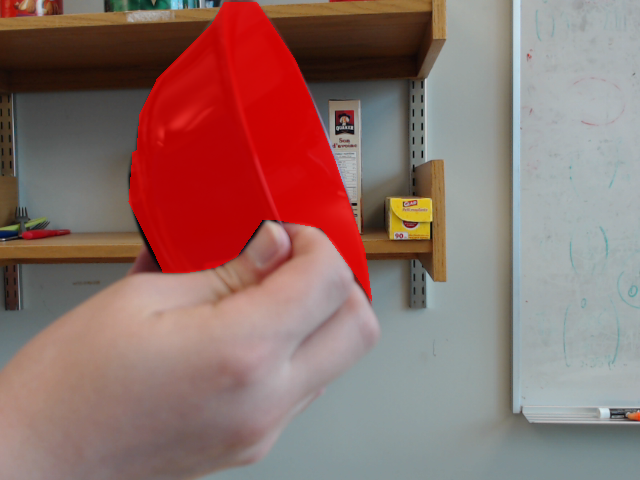}
\end{subfigure}
\begin{subfigure}{.12\textwidth}
    \includegraphics[scale= 0.1]{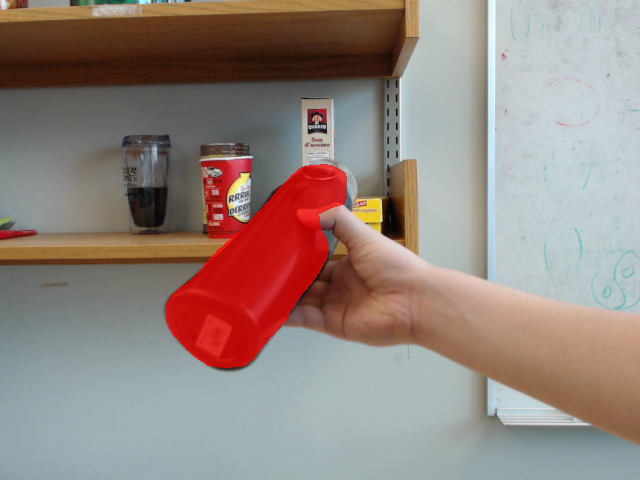}
\end{subfigure}

\begin{subfigure}{.12\textwidth}
    \includegraphics[scale= 0.1]{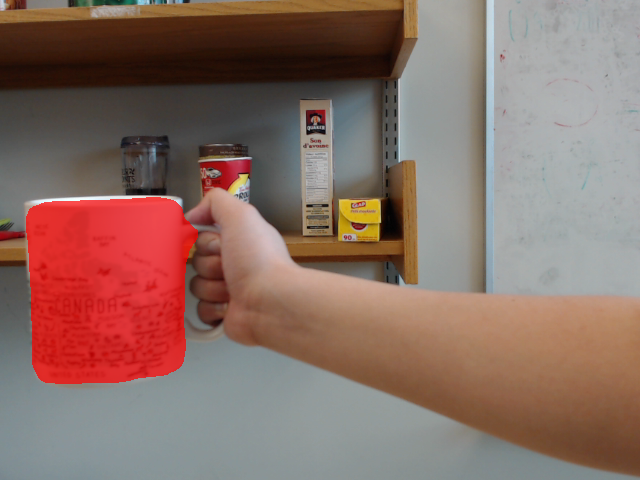}
\end{subfigure}%
\begin{subfigure}{.12\textwidth}
    \includegraphics[scale= 0.1]{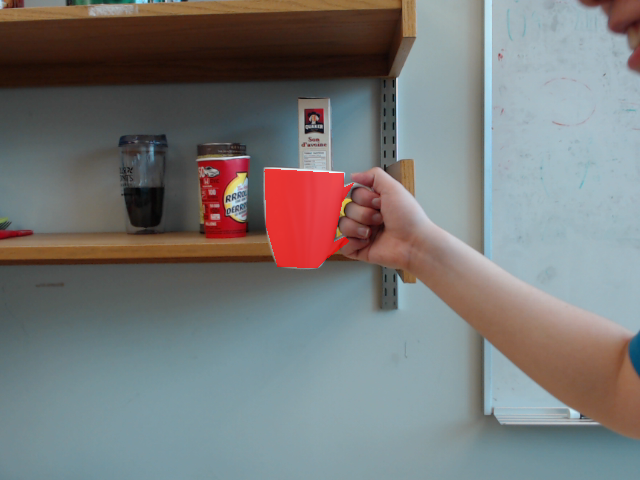}
\end{subfigure}%
\begin{subfigure}{.12\textwidth}
    \includegraphics[scale= 0.1]{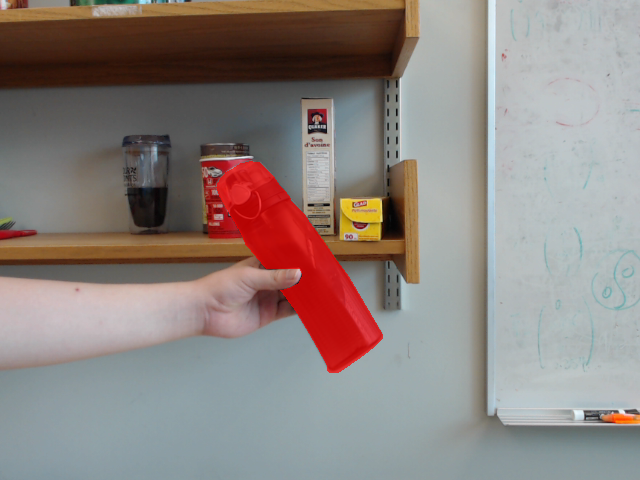}
\end{subfigure}
\begin{subfigure}{.12\textwidth}
    \includegraphics[scale= 0.1]{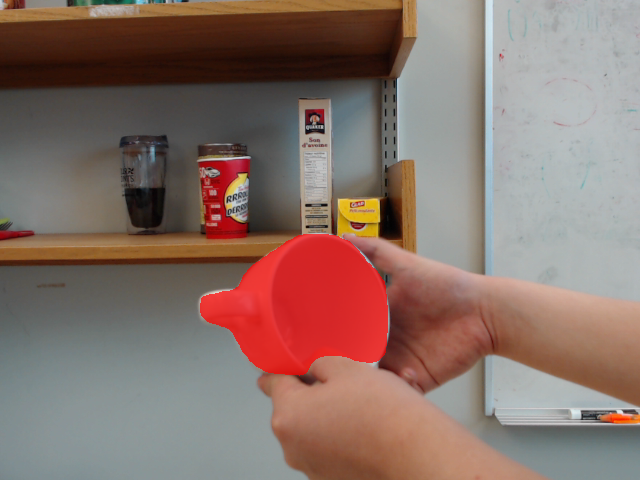}
\end{subfigure}

\begin{subfigure}{.12\textwidth}
    \includegraphics[scale= 0.1]{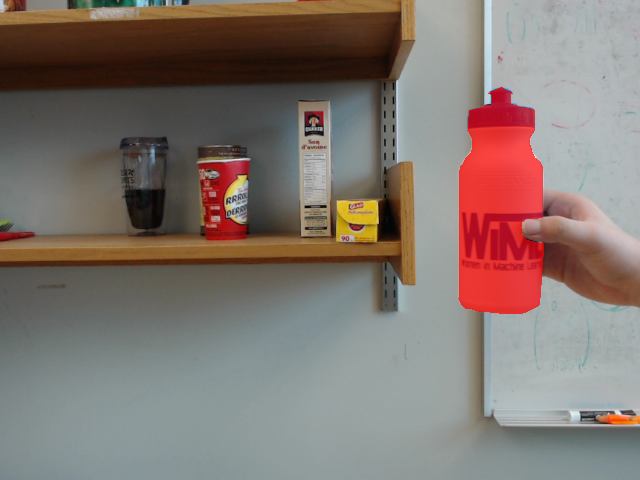}
    \caption{}
\end{subfigure}%
\begin{subfigure}{.12\textwidth}
    \includegraphics[scale= 0.1]{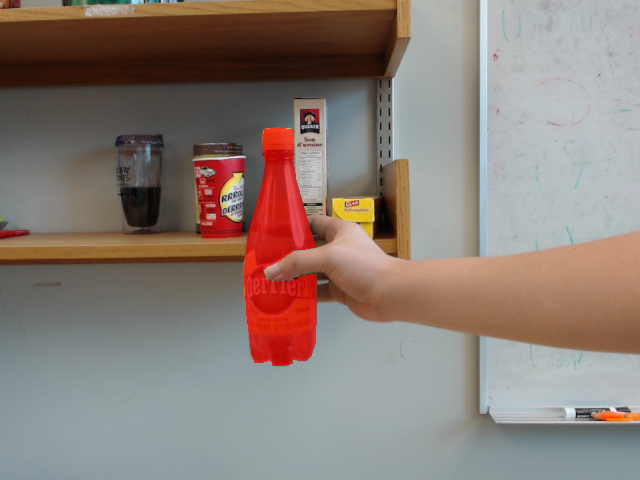}
    \caption{}
\end{subfigure}%
\begin{subfigure}{.12\textwidth}
    \includegraphics[scale= 0.1]{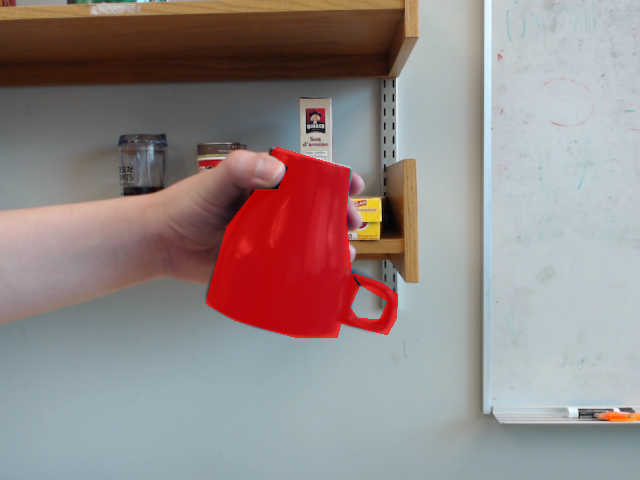}
    \caption{}
\end{subfigure}
\begin{subfigure}{.12\textwidth}
    \includegraphics[scale= 0.1]{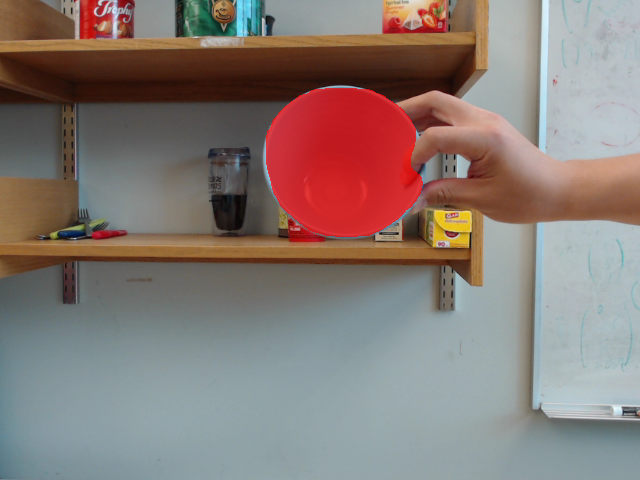}
    \caption{}
\end{subfigure}
\caption{Samples of collected Dataset IVOS, Teaching Objects Setting. (a) Translation split. (b) Scale split. (c) Planar Rotation split. (d) Out-of-plane Rotation. (e) Non rigid transformations.}
\label{fig:ford_objects}
\end{figure*}

\begin{figure*}[ht!]
\centering
\begin{subfigure}{.2\textwidth}
    \includegraphics[scale= 0.17]{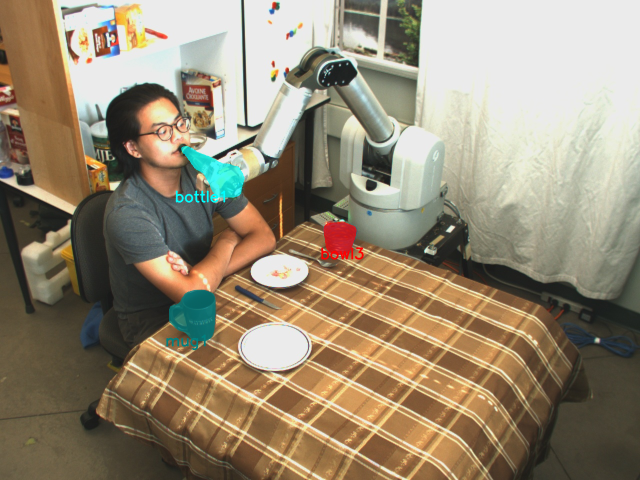}
    \caption{}
\end{subfigure}%
\begin{subfigure}{.2\textwidth}
    \includegraphics[scale= 0.17]{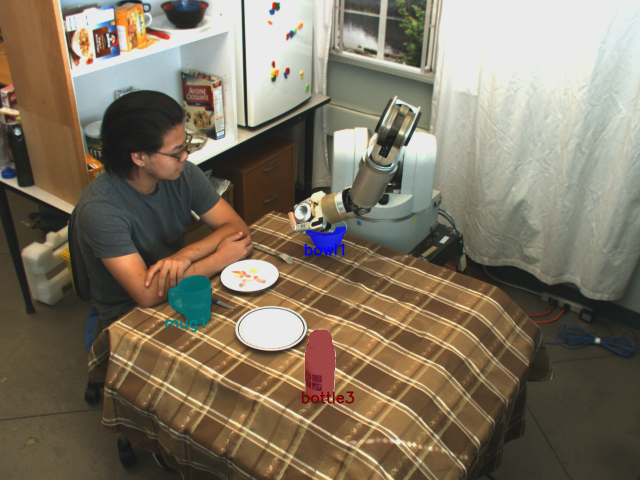}
    \caption{}
\end{subfigure}%
\begin{subfigure}{.2\textwidth}
    \includegraphics[scale= 0.17]{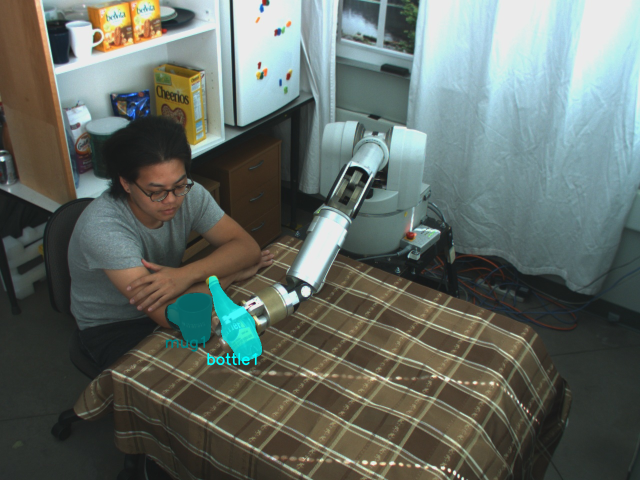}
    \caption{}
\end{subfigure}%
\begin{subfigure}{.2\textwidth}
    \includegraphics[scale= 0.17]{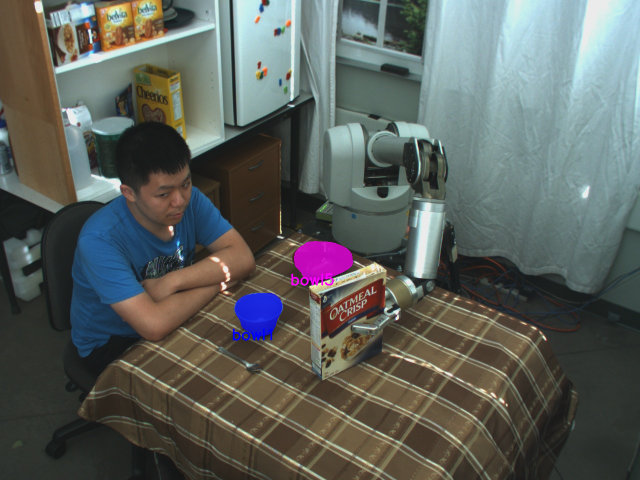}
    \caption{}
\end{subfigure}
\caption{Samples of collected IVOS dataset, Robot manipulation Tasks Setting with segmentation annotation. Manipulation Tasks: (a) Drinking. (b) Stirring. (c) Pouring Water. (d) Pouring Cereal.}
\label{fig:ford_tasks}
\end{figure*}

We collected IVOS for the purpose of benchmarking (I)nteractive (V)ideo (O)bject (S)egmentation in the HRI setting. We collect the dataset in two different settings: (1) Human teaching objects. (2) Manipulation tasks setting. Unlike previous datasets in human robot interaction IVOS dataset provides video sequences for manipulation tasks. In addition to providing segmentation annotation for both teaching videos and manipulation tasks.

\subsection{Human Teaching Objects}
For teaching, videos are collected while a human moves an object with her hand. The unstructured human hand motion naturally provides different views of the object and samples different geometric transformations. We provide transformations such as translation, scale, planar rotation, out-of-plane rotation, and other transformations such as opening the lid of a bottle. Two illumination conditions are provided: day-light and indoor lighting, which sums up to 10 sessions of recording for both illumination and transformations. Figure \ref{fig:ford_objects} shows a sample for the objects being captured under different transformations with the segmentation masks. In each session a video for the object held by a human with relatively cluttered scene background is recorded.

A GRAS-20S4C-C fire-wire camera is used to record the data along with a Kinect sensor \cite{steward2015performance}. The collected data is annotated manually with polygonal masks using the VGG Image Annotator tool \cite{dutta2016via}. The final teaching videos contains 12 object categories, with a total of 36 instances under these categories. The detection crops are provided for all the frames, while the segmentation masks are provided for 20 instances with $\sim$ 18,000 annotated masks.

\subsection{Manipulation Tasks Setting}
The manipulation task benchmark includes two video categories: one with human manipulation, and the other with robot manipulation. Activities of Daily Living (ADL) such as food preparation are the focus for the recorded tasks. The aim of this benchmark is to further improve perception systems in robotics for assisted living. Robot trajectories are created through kinesthetic teaching, and the robot pose way-points are provided in the dataset. In order to create typical robot velocity and acceleration, profiles trajectories were generated from these way-points using splines as is standard in robotics.

The collected sequences are further annotated with segmentation masks similar to the teaching objects setting. Figure \ref{fig:ford_tasks} shows some of the recorded frames with ground-truth annotations. It covers 4 main manipulation tasks: \textit{cutting, pouring, stirring, and drinking} for both robot and human manipulation covering a total of 56 tasks. The dataset contains $\sim$ 8,900 frames with segmentation masks, along with the recorded robot trajectories to enable further research on how to learn these trajectories from visual cues.

\section{Method}
\begin{figure*}[ht!]
	\centering
    \includegraphics[scale= 0.5]{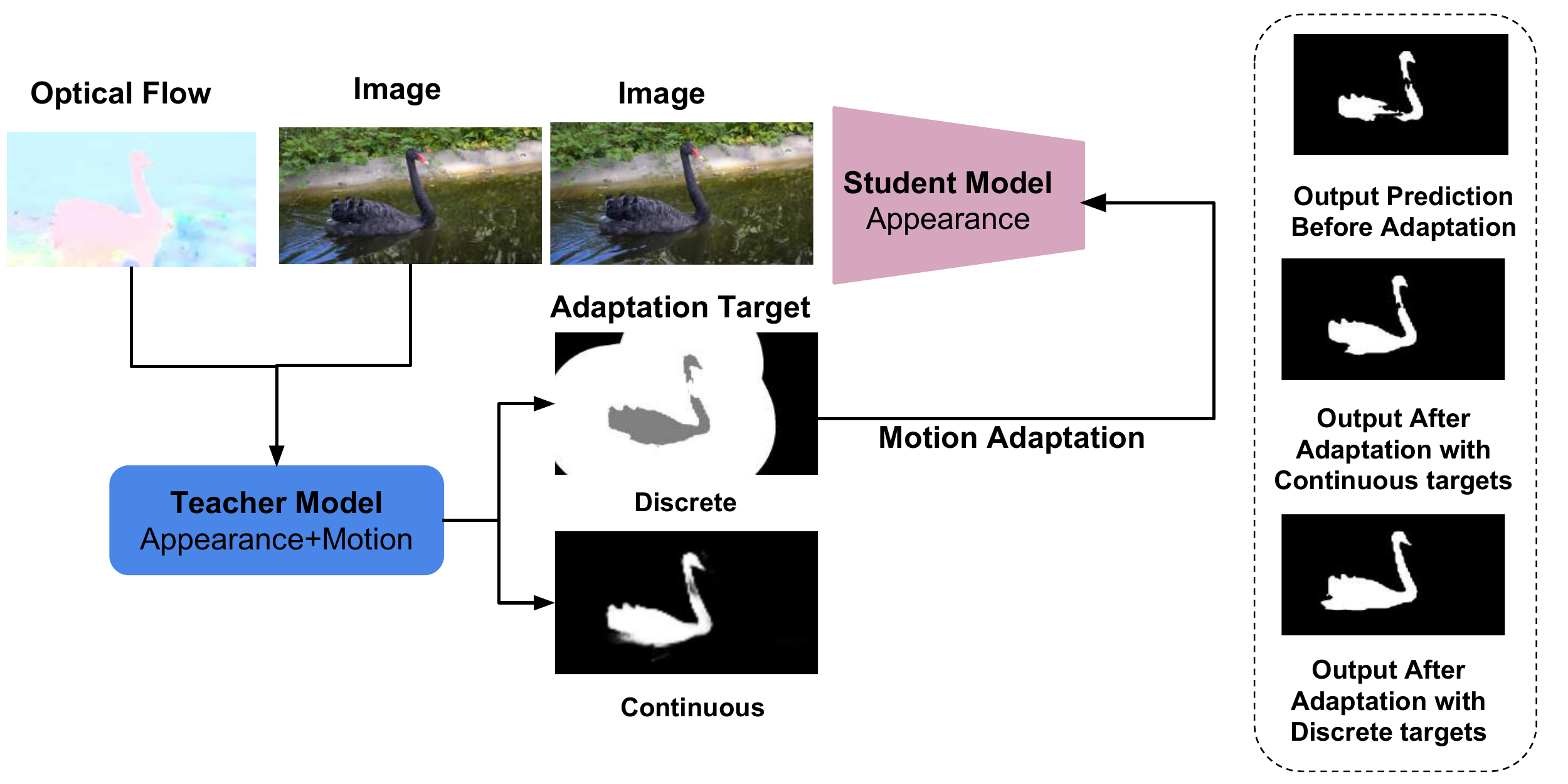}
    \caption{Motion Adaptation of fully convolutional residual networks pipeline.}
    \label{fig:details}
\end{figure*}

\subsection{Baseline Network Architecture}
\label{sec:baseline}

The student model in this work is built on the wide ResNet architecture presented in \cite{wu2016wider}. The network is comprised of 16 residual blocks. Dilated convolution \cite{yu2015multi} is used to increase the receptive field without decreasing the resolution. The output from the network is bilinearly upsampled to the initial image resolution. The loss function used is bootstrapped cross entropy \cite{wu2016bridging}, which helps with class imbalance. It computes the cross entropy loss from a fraction of the hardest pixels. Pre-trained weights on PASCAL dataset for objectness is used from \cite{voigtlaender2017online}, to help the network generalize to different objects in the scene. Then it is trained on DAVIS training set, the student model without adaptation is denoted as the baseline model throughout the paper.

The teacher network incorporates motion from optical flow, where a two-stream wide ResNet for motion and appearance is used. Each stream contains 11 residual blocks for memory efficiency reasons. The output feature maps are combined by multiplying the output activation maps from both motion and appearance streams. After combining features another 5 residual blocks are used with dilated convolution. The input to the motion stream is the optical flow computed using \cite{liu2009beyond}, and converted into RGB representation using the Sintel color wheel representation \cite{baker2011database}.

\subsection{Teacher-Student Adaptation using Pseudo-labels}
There is an analogy between this work and the work in \cite{urban2016deep}, where a student method is learning to mimic a teacher method. In our work the teacher method is a motion dependent one, and the student method tries to mimic the teacher during inference through motion adaptation. The teacher-student training helps the network understands the primary object in the scene in an unsupervised manner. Unlike the work in \cite{voigtlaender2017online} that first fine-tunes the network based on the manual segmentation mask then adapts it on-line with the most confident pixels. Our method provides a natural human robot interaction that does not require manual labelling for initialization.

Our approach provides two different adaptation methods, adapting based on discrete or continuous labels. The teacher network pseudo-labels are initially filtered to remove parts representing the human moving using the output human segmentation from Mask R-CNN \cite{he2017mask}. When discrete labels are used it is based on pseudo-labels from the confident pixels in the teacher network output. Such a method provides superior accuracy, but on the expense of tuning the parameters that determine these confident pixels. Another method that utilizes continuous labels adaptation from the teacher network is also introduced. This method alleviates the need for any hyper-parameter tuning but on the cost of degraded accuracy. Figure \ref{fig:details} summarizes the adaptation scheme, and shows the output pseudo-labels, the output segmentation before and after adaptation.

In the case of discrete pseudo-labels, the output probability maps from the teacher network is further processed in a similar fashion to the semi-supervised method \cite{voigtlaender2017online}. Initially the confident positive pixels are labeled, then a geometric distance transform is computed to label the most confident negative pixels as shown in Algorithm \ref{alg:adapt}.

\begin{algorithm}
    \begin{algorithmic}[1]
    \Function{Teach}{$N$, $X$, $M_{teacher}$, $M_{student}$}
    	\For{\texttt{i in N}}
        \State \texttt{$P_i = M_{teacher}(X_i)$}
        \State \texttt{$\grave{M}_{student}$ = Adapt($P_i$, $M_{student}$)}
      \EndFor
    \EndFunction
    \Statex{\textbf{Discrete Labels Adaptation Method}}
    \Function{Adapt}{$A_t$, $M_{student}$}
    	\State{Mask $\gets$ IGNORED}
        \State{pos\_indices $\gets ( A_{t} >$ POS\_TH )}
        \State{dt $\gets$ \Call{ditance\_transform}{Mask}}
        \State{neg\_indices $\gets ( dt >$ NEG\_DT\_TH )}
        \State{Mask[pos\_indices] $\gets$ 1, Mask[neg\_indices] $\gets$ 0}
        \Return{finetune($M_{student}$,Mask)}
    \EndFunction
    \end{algorithmic} 
    \caption{Motion Adaptation Algorithm. \\ \textbf{Input:} X: images used for teaching. N: number of samples used. $M_{teacher}$: Teacher Model. $M_{student}$: Student Model. \\ \textbf{Output:} $\grave{M}_{student}$: Adapted Student Model.}
    \label{alg:adapt}
\end{algorithm}

\begin{figure}[ht!]
\centering
\begin{subfigure}{.23\textwidth}
    \includegraphics[scale= 0.3]{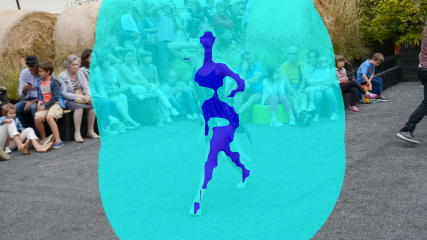}
    \caption{}
\end{subfigure}%
\begin{subfigure}{.23\textwidth}
    \includegraphics[scale= 0.3]{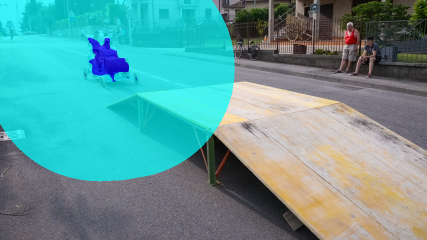}
    \caption{}
\end{subfigure}

\begin{subfigure}{.23\textwidth}
    \includegraphics[scale= 0.3]{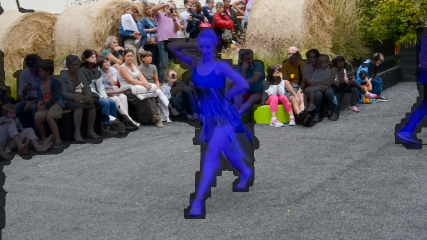}
    \caption{}
\end{subfigure}%
\begin{subfigure}{.23\textwidth}
    \includegraphics[scale= 0.3]{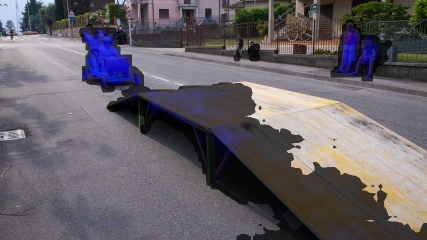}
    \caption{}
\end{subfigure}
\caption{(a,b) Discrete adaptation targets (pseudo-labels), cyan is the unknown region, blue is the confident positive pixels. (c, d) Continuous adaptation targets.}
\label{fig:adapt}
\end{figure}

In the case of continuous labels, the output probability maps are used without further processing. This has the advantage of not using any hyper-parameters or discrete label segmentation. 
It generalizes better to different scenarios on the expense of degraded accuracy. Inspiring from the relation between cross entropy and KL-divergence as in equations \ref{eq:ce_kl}. The cross entropy loss can be viewed as a mean to decrease the divergence between the true distribution $p$ and the predicted one $q$, in addition to the uncertainty implicit in $H(p)$. In our case the true distribution is the probability maps from the teacher network, while the predicted is the student network output. Figure \ref{fig:adapt} shows the difference between the pseudo-labels for both discrete and continuous variants. Conditional random fields is used as a post-processing step on DAVIS and FBMS.

\begin{subequations}
\begin{equation}
D_{KL}(p|q) = \sum_i{p_i \log{\frac{p_i}{q_i}}}
\end{equation}
\begin{equation}
D_{KL}(p|q) = \sum_i{p_i \log{\frac{1}{q_i}}} - H(p)
\end{equation}
\begin{equation}
H(p, q) = H(p) + D_{KL}(p|q)
\end{equation}
\label{eq:ce_kl}
\end{subequations}

\section{Experimental Results}
\subsection{Experimental Setup}
For all experiments the DAVIS training data is used to train our Appearance model and the Appearance+Motion model. The optimization method used is Adam \cite{kingma2014adam} with learning rate $10^{-6}$ during training, and $10^{-5}$ during on-line adaptation. In on-line adaptation 15 iterations are used in the scale/rotation experiments and 50 in the tasks experiments. Adaptation is only conducted once at the initialization of the video object segmentation. The positive threshold used to identify highly confident positive samples is 0.8, and the negative threshold distance to the foreground mask is 220 in case of DAVIS benchmark. Since IVOS is recorded in an indoor setup, a negative distance threshold of 20 is used.

\begin{table*}[ht!]
\centering
\caption{Quantitative comparison on DAVIS benchmark. MotAdapt-1: Continuous Labels, MotAdapt-2: Discrete Labels. }
\label{table:soa}
\begin{tabular}{|l|l|l|l|l|l|l|l|l|l|l|}
\hline
\multicolumn{2}{|l|}{Measure} & NLC\cite{faktor2014video} & SFL\cite{cheng2017segflow} & LMP \cite{tokmakov2016learning} & FSeg \cite{jain2017fusionseg} & LVO \cite{tokmakov2017learning} & ARP \cite{kohprimary} & Baseline & MOTAdapt-1 & MOTAdapt-2 \\ \hline
\multirow{3}{*}{$\mathcal{J}$} & Mean & 55.1 & 67.4 & 70.0 & 70.7 & 75.9 & 76.2 & 74.0 & 75.3 & \textbf{77.2} \\ 
                   & Recall & 55.8 & 81.4 & 85.0 & 83.5 & 89.1 & \textbf{91.1} & 85.7 & 87.1 & 87.8 \\ 
                   & Decay & 12.6 & 6.2 & 1.3 & 1.5 & \textbf{0.0} & \textbf{0.0} & 7.0 & 5.0 & 5.0 \\ \hline
\multirow{3}{*}{$\mathcal{F}$} & Mean & 52.3 & 66.7 & 65.9 & 65.3 & 72.1 & 70.6 & 74.4 & 75.3 & \textbf{77.4} \\ 
                   & Recall & 51.9 & 77.1 & 79.2 & 73.8 & 83.4 & 83.5 & 81.6 & 83.8 & \textbf{84.4} \\ 
                   & Decay & 11.4 & 5.1 & 2.5 & 1.8 & 1.3 & 7.9 & \textbf{0.0} & 3.3 & 3.3 \\ \hline
\end{tabular}
\end{table*}

\begin{table*}[ht!]
\centering
\caption{Quantitative results on FBMS dataset (test set).}
\label{table:fbms}
\begin{tabular}{|l|c|c|c|c|c|c|c|}
\hline
Measure & FST \cite{papazoglou2013fast} & CVOS \cite{taylor2015causal} & CUT \cite{keuper2015motion} & MPNet-V\cite{tokmakov2016learning}  & LVO\cite{tokmakov2017learning} & Base & ours  \\ \hline
$\mathcal{P}$ & 76.3 & 83.4 & 83.1 & 81.4 & \textbf{92.1} & 80.8 & 80.7 \\
$\mathcal{R}$ & 63.3 & 67.9 & 71.5 & 73.9 & 67.4 & 76.1 & \textbf{77.4} \\
$\mathcal{F}$ & 69.2 & 74.9 & 76.8 & 77.5 & 77.8 & 78.4 & \textbf{79.0} \\ \hline
\end{tabular}
\end{table*}

\begin{table*}[]
\caption{mIoU on IVOS over the different transformations and tasks. IVOS dataset teaching is conducted on few samples from the translation, then evaluating on scale, rotation and manipulation tasks. MotAdapt-1: Continuous Labels. MotAdapt-2: Discrete Labels.}
\centering
\begin{tabular}{|l|c|c|c|}
\hline
Model & Scale & Rotation & Manipulation Tasks \\ \hline
Baseline & 14.5 & 13.8 & 14.7 \\ 
MotAdapt-1 & 63.8 & 49.5 & 30.2 \\ 
Mot-Adapt-2 & \textbf{69.0} & \textbf{51.5} & \textbf{40.6} \\ \hline
\end{tabular}
\label{table:fordsm}
\end{table*}

\begin{figure*}[ht!]
\centering
\begin{subfigure}{.22\textwidth}
    \makebox[20pt]{\raisebox{30pt}{\rotatebox[origin=c]{90}{LVO}}}%
    \includegraphics[scale= 0.96]{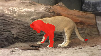}
\end{subfigure}%
\begin{subfigure}{.13\textwidth}
    \includegraphics[scale= 0.65]{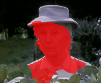}
\end{subfigure}%
\begin{subfigure}{.13\textwidth}
    \includegraphics[scale= 0.695]{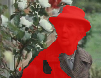}
\end{subfigure}%
\begin{subfigure}{.18\textwidth}
    \includegraphics[scale= 0.72]{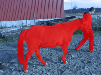}
\end{subfigure}

\begin{subfigure}{.22\textwidth}
	\makebox[20pt]{\raisebox{30pt}{\rotatebox[origin=c]{90}{MotAdapt}}}%
    \includegraphics[scale= 0.96]{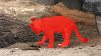}
\end{subfigure}%
\begin{subfigure}{.13\textwidth}
    \includegraphics[scale= 0.65]{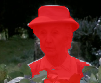}
\end{subfigure}%
\begin{subfigure}{.13\textwidth}
    \includegraphics[scale= 0.695]{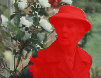}
\end{subfigure}%
\begin{subfigure}{.18\textwidth}
    \includegraphics[scale= 0.72]{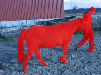}
\end{subfigure}

\caption{Qualitative Evaluation on the FBMS dataset. Top: LVO \cite{tokmakov2017learning}. Bottom: ours.}
\label{fig:fbms}
\end{figure*}

\begin{figure*}[ht!]
\centering
\begin{subfigure}{.19\textwidth}
    \includegraphics[scale= 0.45]{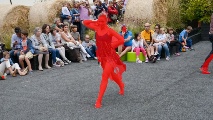}
\end{subfigure}%
\begin{subfigure}{.19\textwidth}
    \includegraphics[scale= 0.45]{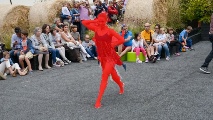}
\end{subfigure}%
\begin{subfigure}{.19\textwidth}
    \includegraphics[scale= 0.45]{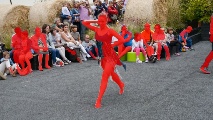}
\end{subfigure}%
\begin{subfigure}{.19\textwidth}
    \includegraphics[scale= 0.45]{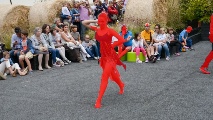}
\end{subfigure}

\begin{subfigure}{.19\textwidth}
    \includegraphics[scale= 0.45]{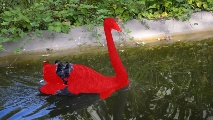}
    \caption{}
\end{subfigure}%
\begin{subfigure}{.19\textwidth}
    \includegraphics[scale= 0.45]{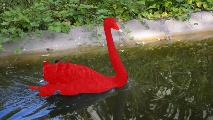}
    \caption{}
\end{subfigure}%
\begin{subfigure}{.19\textwidth}
    \includegraphics[scale= 0.45]{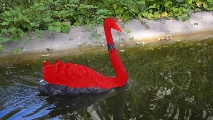}
    \caption{}
\end{subfigure}%
\begin{subfigure}{.19\textwidth}
    \includegraphics[scale= 0.45]{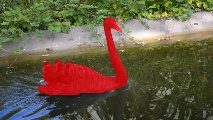}
    \caption{}
\end{subfigure}
\caption{Qualitative evaluation on DAVIS16. (a) LVO \cite{tokmakov2017learning}. (b) ARP \cite{kohprimary}. (c) Baseline. (d) MotAdapt.}
\label{fig:davis}
\end{figure*} 

\begin{figure*}[ht!]
\centering
\begin{subfigure}{.18\textwidth}
    \includegraphics[scale= 0.6]{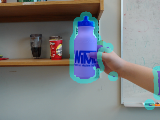}
    \caption{}
\end{subfigure}%
\begin{subfigure}{.18\textwidth}
    \includegraphics[scale= 0.6]{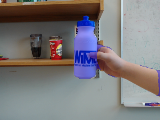}
    \caption{}
\end{subfigure}%
\begin{subfigure}{.18\textwidth}
    \includegraphics[scale= 0.3]{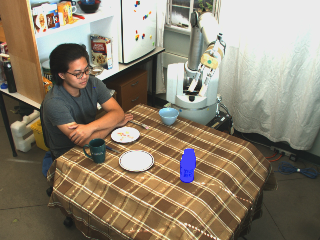}
    \caption{}
\end{subfigure}%
\begin{subfigure}{.18\textwidth}
    \includegraphics[scale= 0.3]{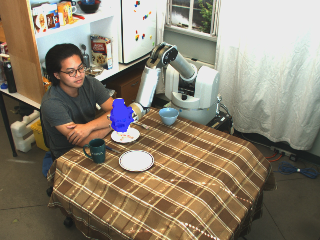}
    \caption{}
\end{subfigure}

\caption{Qualitative evaluation on IVOS Manipulation Tasks Setting. (a) Teaching Phase, Discrete Labels. (b) Teaching Phase, Continuous Labels. (c) Inference Phase before manipulation. (d) Inference Phase, during manipulation.}
\label{fig:fordsm_qual}
\end{figure*} 

\subsection{Generic Video Object Segmentation}
In order to evaluate the performance of our proposed motion adaptation (MotAdapt) method with respect to the state-of-the-art, we experiment on generic video object segmentation datasets. Table \ref{table:soa} shows quantitative analysis on DAVIS benchmark compared to the state-of-the-art unsupervised methods. One of the variants of MotAdapt based on discrete labels outperforms the state of the art with 6.8\% in F-measure, and 1\% in mIoU. Table \ref{table:fbms} shows quantitative results on FBMS dataset, where our MotAdapt outperforms the state of the art with 1.2\% in F-measure and 10\% in recall. 

Figure \ref{fig:fbms} shows qualitative results on FBMS highlighting the improvement gain from motion adaptation compared to LVO \cite{tokmakov2017learning}. Figure \ref{fig:davis} shows qualitative evaluation on DAVIS, where it demonstrates the benefit from motion adaptation compared to the baseline (top row), and compared to LVO \cite{tokmakov2017learning} and ARP \cite{kohprimary} (bottom row).

\subsection{Video Object Segmentation in HRI Setting}
Our method is evaluated in the HRI scenario on our dataset IVOS. The teaching is performed on the translation sequences, with only the first two frames used to generate pseudo-labels for adaptation. An initial evaluation is conducted on both scale and rotation sequences, in order to assess the adaptation capability to generalize to different poses and transformations. Table \ref{table:fordsm} shows the comparison between the baseline method without adaptation, and the two variants of motion adaptation on the scale, rotation and tasks sequences. The discrete and continuous variants for our motion adaptation outperform the baseline with 54.5\% and 49.3\% respectively on the scale sequences. Similarly on the rotation sequences it outperforms the baseline with 37.7\% and 35.7\% respectively. The main reason for this large gap, is that general segmentation methods will segment all objects in the scene as foreground, while our teaching method adaptively learns the object of interest that was demonstrated by the human teacher. 

All manipulation tasks sequences where the category bottle existed is evaluated and cropped to include the working area. Our method outperforms the baseline on the tasks with 25.9\%. The first variant of our adaptation method generally outperforms the second variant with continuous labels adaptation. However the second variant has the advantage that it can work on any setting such as DAVIS and IVOS without tuning any hyper-parameters. Figure \ref{fig:fordsm_qual} shows the output from our adaptation method when it is recognized by the robot, and while the robot has successfully manipulated that object.

\section{Conclusions}
In this paper we proposed a novel approach for visual learning by instruction. Our proposed motion adaptation (MotAdapt) method provides a natural interface to teaching robots to segment novel object instances. This enables robots to manipulate and grasp these objects. Two variants of the adaptation scheme is experimented with. Our results show that Mot-Adapt outperforms the state of the art on DAVIS and FBMS and outperforms the baseline on IVOS dataset.

{\small
\bibliographystyle{IEEEtranS}
\bibliography{IEEEfull}
}

\end{document}